\documentclass[10pt,twocolumn,letterpaper]{article}

\usepackage{cvpr}              % To produce the CAMERA-READY version
\usepackage{graphicx}
\usepackage{amsmath}
\usepackage{amssymb}
\usepackage{booktabs}

\usepackage[pagebackref,breaklinks,colorlinks]{hyperref}

\usepackage[capitalize]{cleveref}
\crefname{section}{Sec.}{Secs.}
\Crefname{section}{Section}{Sections}
\Crefname{table}{Table}{Tables}
\crefname{table}{Tab.}{Tabs.}

\def\cvprPaperID{*****} % *** Enter the CVPR Paper ID here
\def\confName{CVPR}
\def\confYear{2023}

\begin{document}

\title{Overlooked Video Classification in Weakly Supervised Video Anomaly Detection}

\author{Weijun Tan\\
LinkSprite Technologies, USA\\ 
and Jovisin-Deepcam Research, Shenzhen, China\\
{\tt\small weijun.tan@linksprite.com}\\
{\tt\small sz.twj@jovision.com}
\and
Qi Yao, and Jingfeng Liu\\
Jovision-Deepcam Research\\
Shenzhen, China\\
{\tt\small \{sz.yaoqi,sz.ljf\}@jovision.com} \\
{\tt\small \{qi.yao,jingfeng.liu\}@deepcam.com}
}
\maketitle

%%%%%%%%% ABSTRACT
\begin{abstract}
Current weakly supervised video anomaly detection algorithms mostly use multiple instance learning (MIL) or their varieties. Almost all recent approaches focus on how to select the correct snippets for training to improve performance. They overlook or do not realize the power of video classification in boosting the performance of anomaly detection. In this paper, we study the power of video classification supervision explicitly using a BERT or LSTM. With this BERT or LSTM, CNN features of all snippets of a video can be aggregated into a single feature which can be used for video classification. This simple yet powerful video classification supervision, combined with the MIL and RTFM framework, brings extraordinary performance improvement on all three major video anomaly detection datasets. Particularly it improves the mean average precision (mAP) on the XD-Violence from SOTA 78.84\% to new 82.10\%. These results demonstrate this video classification can be combined with other anomaly detection algorithms to achieve better performance. The code is publicly available at xxx.       

\end{abstract}

%%%%%%%%% BODY TEXT
\section{Introduction}
\label{sec:intro}

Surveillance cameras are widely used in public places for safety purposes.  Enpowered by machine learning and artificial intelligence, surveillance cameras become smarter using automatic object or event detection and recognition.  Video anomaly detection is to identify the time and space of abnormal objects or events in videos. Examples include industrial anomaly detection and security anomaly detection, and more. 

Depending on the annotation of the training data and the algorithms, anomaly detection is categorized into three types - unsupervised, supervised, and weakly supervised. The unsupervised one learns only on normal videos assuming the unseen anomalous videos have high reconstruction errors. This approach's performance is usually poor because it lacks  knowledge of the abnormality in anomalous videos and inability to learn the normal patterns in normal videos. The supervised one is expect to have the best performance. However, due to the fact that the frame-level annotation is very time consuming to get and prone to human mistakes, it is less studied. In weakly supervised one, since only video level annotation of if there is anomaly in a video is needed, the dataset is a lot easier to get and robust to human mistakes. It draws most attentions in the video anomaly detection area.  

In the weakly supervised anomaly detection, a multiple instance learning (MIL) or its variety is typically used \cite{UCF-Crime}.  From a pair of abnormal and normal videos, a positive bag of instances is formed on the abnormal video, and a negative bag of instances on the normal video. A pretrained CNN network is used to extract a feature on a snippet of video frames. A classification network is trained on all the instances of these two bags. The one instance with the maximum classification score in a bag is chosen to represent the bag. The MIL tries to maximize the separation between the maximum scores of the positive bag and the negative bag. 

In almost all follow-up studies, different approaches are proposed on how to select the best quality snippets to train the model. Some choose multiple snippets instead of one out of a video \cite{RTFM}, others choose a sequence of consecutive snippets \cite{AAAI22}, \cite{MIST}. Some of them use the snippet classification score to choose snippets, others use other metrics including feature magnitude \cite{RTFM}. Some use GCN to improve the quality of the chosen snippets \cite{adgcn_cvpr19}.  

However, almost all of them overlook or do not fully realize the power of the video classification and its impact on the anomaly detection performance. In anomaly detection, the videos are classified to anomalous or normal videos. This strong information has been overlooked except in RTFM \cite{RTFM}, \cite{AAAI22}, and \cite{XD-Violence}. In RTFM, the top-k snippets with maximum feature magnitude are chosen per video, and the mean of their classification scores is used as a video classification score in the binary cross entropy (BCE) loss, even though the authors do not call it so.

In \cite{XD-Violence}, a GCN is used to approximately model the video classification, and a video classification BCE loss is used. The work that is most relevant to ours is \cite{AAAI22}. While we are studying an explicit video classification using BERT \cite{BERT-0}, \cite{BERT}, we find that they use a transformer to model the video classification with a BCE loss. In addition to this video classification, the transformer is also used to refine the CNN feature . They propose a multiple sequence learning (MSL) finding consecutive snippets to improve the training, which is claimed as their main contribution. However, in our work we find that a BERT or a transformer does not necessarily fulfill both tasks of video classification and feature refinement at the same time. We find that it does not help the feature refinement, so we solely study its role in video classification. With this simple single change, without MSL or RTFM, we achieve superior performance on the UCF-Crime \cite{UCF-Crime} and the ShanghaiTech \cite{shanghaitech} datasets.  

We go further to use this BERT video classification on top of the RTFM. We combine their BCE loss and our proposed BERT-enabled BCE loss, and achieve extraordinary performance on the XD-Violence dataset. Based on these results, we demonstrate the power of the video classification supervision in anomaly detection. It can work alone or combine other techniques like RTFM to boost the performance of anomaly detection.  

Our contributions are summarized as follows, 
\begin{itemize}
\item{We explicitly study the power of video classification supervision in weakly supervised video anomaly detection. This video classification is achieved with a BERT on snippet CNN features. We find that the BERT should only be used for the video classification, but should not be used for feature refinement. \textbf{It is very important to emphasize that the combination of two existing ideas (BERT and MIL) should not be taken as our main contribution. Instead, our key contribution is we find out the fact that the power of video classification has been previously overlooked and now the gap is filled in this work. As an ablation study, we implement a simpler LSTM based video classifier. Even though its complexity is a lot lower, its performance is almost the same as the BERT. }} 

\item{There are two inference modes of this proposed scheme. The second online mode offers a very attractive low complexity option, even though it only gets partial performance improvement from the video classification supervision.} 
\item{We study this algorithm alone in the standard MIL framework on the UCF-Crime and the ShanghaiTech datasets. We test RGB, Flow or RGB+Flow modality. This simple introduction of video classification in anomaly detection brings superior performance improvement on every modality. On the RGB+flow modality, we achieve the best ROC-AUC performance, exceeding the SOTA by 1.5\%.}  
\item{We study this algorithm on top of the RTFM \cite{RTFM} on the UCF-Crime and the XD-Violence datasets. We test the RGB modality only.  While our algorithm only achieves a marginal ROC-AUC performance improvement on the UCF-Crime dataset, it achieve nearly 3.51\% AP performance improvement on the XD-Violence dataset. \textbf{This improvement demonstrates that our proposed explicit video classification can combine with many other video anomaly detection algorithms where an explicit video classification is not used. } } 
\end{itemize}

\section{Related Work}

Unsupervised anomaly detection assume only normal training data is available and solves this problem with one-class classification using hand-crafted features or deep learning features. Typical approaches use pre-trained CNN, apply constraints on the latent space of normal manifold to learn normality representation, or use data reconstruction error with generative models. There are very few work on the supervised learning for anomaly detection since the frame level annotation is very hard to get. Two examples are \cite{MMM19} and \cite{localizanomaly}. For a review of video anomaly detection, the readers are referred to \cite{jimaging4020036} and \cite{anomaly-review}. 

Weakly supervised anomaly detection has shown substantially improved performance over the self supervised approaches by leveraging the available video-level annotations. These annotation only gives a binary label of abnormal or normal for a video. Sultani et al. \cite{UCF-Crime} propose the MIL framework using only video-level labels and introduce the large-scale anomaly detection dataset, UCF-Crime. This work inspires quite a few follow-up studies \cite{adgcn_cvpr19}, \cite{wsal_tip21}, \cite{STAD},  \cite{CRFD}, \cite{IJCAI21}, \cite{MIST}, \cite{RTFM}, \cite{AAAI22}. 

However, in the MIL-based methods, abnormal video labels are not easy to be used effectively. Typically, the classification score is used to tell if a snippet is abnormal or normal. This score is noisy in the positive bag, where a normal snippet can be mistakenly taken as the top abnormal event in an anomaly video. To deal with this problem, Zhong et al. \cite{adgcn_cvpr19} treat this problem as a binary classification under noisy label problem and use a graph convolution neural (GCN) network to clear the label noise.  In \cite{MIST}, a multiple instance self-training framework (MIST) is proposed to  efficiently refine task-specific discriminative representations with a multiple instance pseudo label generator and a self-guided attention boosted feature encoder. In \cite{CRFD}, a weakly-supervised spatio-temporal anomaly detection is proposed to localize a spatio-temporal tube that encloses the abnormal event. In \cite{CRFD}, causal temporal cue and feature discrimination are explored. In \cite{wsal_tip21}, a high-order context encoding model is used to to encode temporal variations as well as high-level semantic information for weak-supervised anomaly detection. 

In RTFM \cite{RTFM}, a robust temporal feature magnitude (RTFM) is used to select the most reliable abnormal snippets from the abnormal videos and the normal videos.  They unify the representation learning and anomaly score learning by an temporal feature ranking loss, enabling better separation between normal and abnormal feature representations, improving the exploration of weak labels compared to previous MIL methods. In \cite{AAAI22} a multiple sequence learning (MSL) is used. The MSL uses a sequence of multiple instances as the optimization unit instead of one single instance in the MIL. In addition, a transformer is used to refine the snippet features. A video classification is used with the transformer classification token. 

In \cite{XD-Violence}, both video and audio signals are used to detect anomaly in video with audio. They use a GCN to model the long-term and local dependencies. At the same time, they release so far the largest video anomaly detection dataset - the XD-Violence dataset.   

\begin{figure}[t]
    \centering
    \includegraphics[scale=0.4]{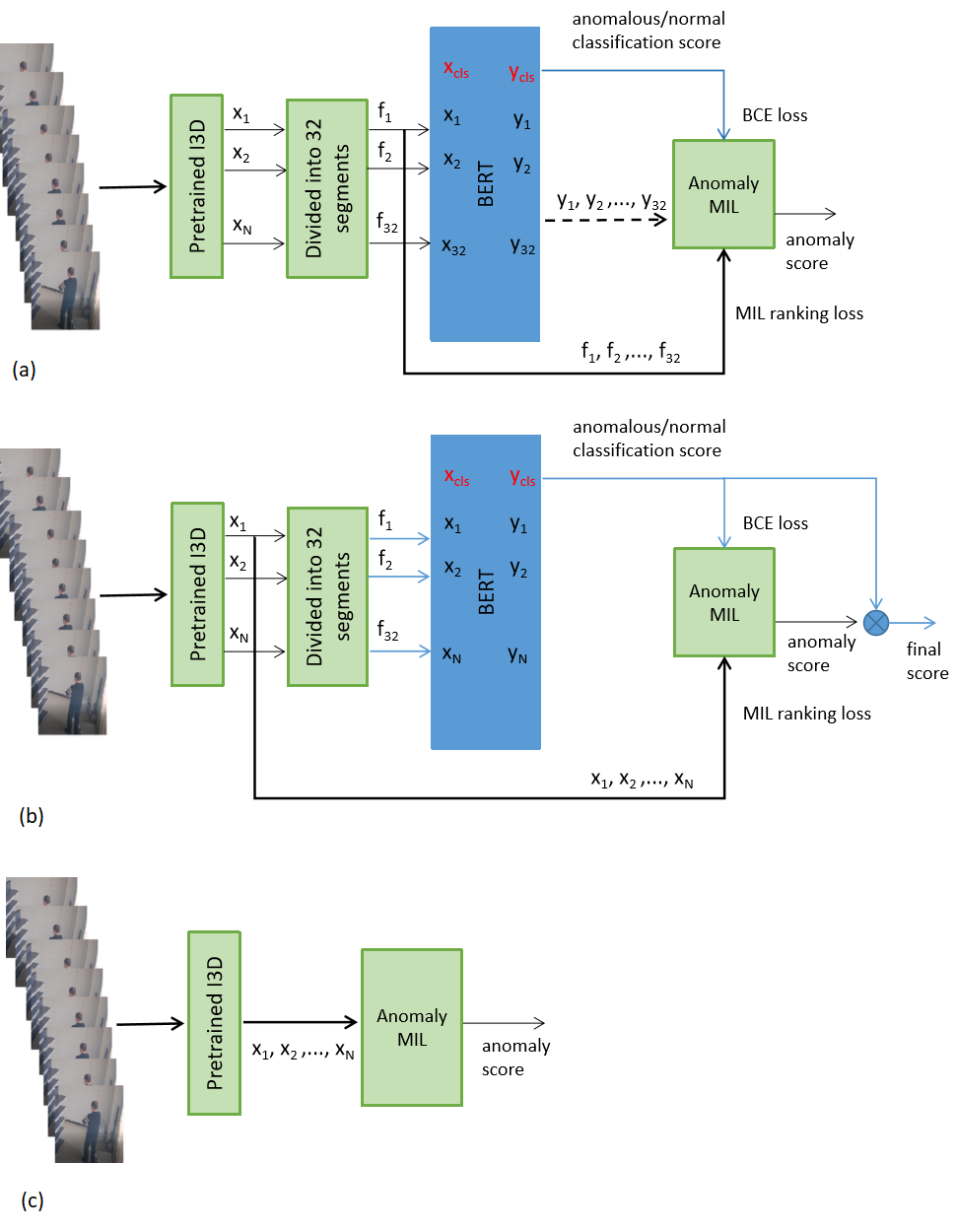}
    \caption{Block diagram of our anomaly detection with BERT video classification, (a) training, (b) testing with video classification, (c) testing without video classification.} 
    \label{fig1}
\end{figure}

%------------------------------------------------------------------------
\section{Proposed Methods}

We propose to use the BERT as a video classifier for it extraordinary capability to aggregate information with both spatial and temporal attention. The diagrams of the training and testing pipelines are shown in Figure 1. More details will be given below. 

Let's first define some terms which may be confusing. Video classification refers to classify every video to be normal (negative) or anomalous (or abnormal, positive). Snippet is defined as a sequence of video frames of fixed length, in this work, 16. Segment is defined as a sequence of snippets. In this work we follow previous work to divide every training and validation videos to equal length 32 segments. In testing video, either snippet or segment can be used.  

\subsection{Introduction to BERT}

The transformer first appeared in 2017 in a paper entitled “Attention Is All You Need” \cite{attention}. It is a very successful Natural Language Processing (NLP) model and has become one of the break-through innovations in recent years. Since then, the transformer has been extended to almost every field of machine learning, including image classification \cite{ViT}, object detection \cite{DETR}, and video understanding \cite{video_transformer} and many more. The transformer pay attention to every element of a sequence of input data in self-attention and extracts the traces of the entire data set.

Following the success of the transformer - a one directional model, the Bi-directional Encoder Representations from Transformers (BERT) \cite{BERT-0} is a bidirectional self-attention model, which has also been a big success in many downstream NLP tasks. The bidirectional property enables BERT to fuse the contextual information from both directions. Moreover, BERT introduces challenging unsupervised pre-training tasks which leads to useful representations for many tasks. BERT is introduced in \cite{BERT} for video action recognition and achieves SOTA performance on two major action recognition datasets, UCF-101 \cite{UCF101} and JHMDB-51 \cite{JHMDB}. We are inspired by the BERT, particularly its application on action recognition \cite{BERT}.    

In \cite{BERT}, the BERT is used as a late pooling function to replace the previous widely used global average pooling (GAP). The input to the BERT is the internal CNN feature map usually taken before the GAP and FC layers. To preserve the positional
information, a learned positional encoding is added to the extracted features. In order to perform classification, additional classification token $x_{cls}$ is appended as in \cite{BERT-0}. The classification is implemented with the corresponding classification vector $y_{cls}$, which is sent to FC layers for classification prediction. 

The general single head self-attention model of BERT is formulated as:

\begin{equation}
    y_i = PFFN \left( \frac{1}{N(x)} \sum_j g(x_j) f(x_i,x_j)\right)
    \label{eq1}
\end{equation}
where $x_i$ values are the input input vector including positional encoding; $i$ indicates the index of the target output temporal position; $j$ denotes all possible combinations; and $N(x)$ is a normalization term. The function $g()$ is the linear projection inside the self-attention mechanism of BERT, whereas function $f()$ denotes the similarity between $x_i$ and $x_j$ as $f(x_i,x_j) = softmax_j (\theta(x_i)^T \phi(x_j))$, where the functions $\theta(·)$ and $\phi(·)$ are linear projections. The learnable functions $g(·)$, $\theta(·)$ and $\phi(·)$ try to project the feature embedding vectors to a better space where the attention mechanism works more efficiently. The outputs of $g(·)$, $\theta(·)$ and $\phi(·)$ functions are usually called value, query and key \cite{attention} . $PFFN$ is position-wise feed-forward network (PFFN) applied to all positions separately and identically as  $PFFN(x) = W_2GELU(W_1x + b_1) + b_2$, where $GELU(·)$ is the Gaussian Error Linear Unit (GELU) activation function \cite{attention}. The classification vector $y_{cls}$ takes similar form as for $y_i$, 

\begin{equation}
    y_{cls} = PFFN \left( \frac{1}{N(x)} \sum_j g(x_j) f(x_{cls},x_j)\right)
    \label{eq2}
\end{equation}

In our work, since our focus is to study the impact of the video classification, our main goal is to use the learned classification embedding $y_{cls}$ which aggregates the temporal features of a video into a single feature for video classification. The input vector $x_i$'s are the feature vectors of the segments. The learned better subspace for feature representation ${y_i}$'s are a plus if they work better than the original features $x_i$'s in anomaly MIL. However, we have a strong motivation why we do not have to use it for application purpose - to have a solution in inference mode that has low complexity as the original MIL framework \cite{UCF-Crime}.  

\subsection{Proposed Training Process}

Shown in Figure 1 are the block diagram of our proposed anomaly detection with explicit video classification. We call this proposed solution MIL-BERT.  

Figure 1(a) is the training pipeline. First, given a video, frames are extracted into snippets of 16 frames. A pretrained 3D CNN backbone network is used to extract CNN feature. In the diagram we demonstrate a I3D network \cite{I3D}, but other networks, C3D \cite{C3D} and newer X3D \cite{X3D}, or MoViNet \cite{MoViNet} can also be used. This backbone network keeps frozen and does not participate our training. The output snippet features are denotes as $f_i, j=1,2,...,N$, where N is the number of snippets of the video. These features are divided equally into 32 segments $x_i,i=1,2,...,32$. This function is defined as, 

\begin{equation}
    x_i = seg(\{f_j,j=1,2,...,N\}) 
    \label{eq3}
\end{equation}
where $seg()$ stands for the segmentation on snippets. Its inverse function is denoted as $seg^{-1}()$.

The segmented features $x_i$ are sent to the BERT as input temporal features. The output features $y_i$ in a different sub-space and the classification feature $y_{cls}$ are output after all the bidirectional attention mechanisms in BERT. The standard MIL anomaly detection framework can take the feature $x_i$ or $y_i$ as input. We propose to send the video classification $\hat{y}$ out of the BERT to the MIL as an input. This similar idea is also used in \cite{AAAI22}. 

We keep using the MIL ranking loss function with the smoothness and sparsity term as in \cite{UCF-Crime}. We add the video classification binary cross entropy (BCE) loss onto it. So the overall loss function is defined as, 

\begin{align}
    l = & \max(0, 1-\max\limits_{i \in B_a} s(v_a^i) + \max\limits_{i \in B_n} s(v_n^i))\nonumber \\
        & -y_alog(p(\hat{y}_{cls,a}))-(1-y_n)log(1-p(\hat{y}_{cls,n}))
    \label{eq4}
\end{align}
where the subscript $a$ and ${n}$ denotes anomalous and normal video, $v$ is an input feature instance, which can be $x_i$, ${y_i}$, $f_i$ in our work. $B_a$ and $B_n$ are the bags of segments in the abnormal and normal video, $s(.)$ is the predicted anomaly scoring function in range of 0 and 1. The function $max$ is taken over all instances in a bag. It is expected that in the positive bag, the highest-scored instance is a true abnormal segment. The highest-scored instance in the negative bag is the one most similar to the positive bag, but is actually a negative instance. This makes the negative instance a hard one and therefore benefits the discriminability in the model training. In the BCE loss part, we only keep part of the standard form since $y_a=1, y_n=0$, $p$ is the scoring function of the video classifier. The smoothness and sparsity terms are still used, even though they are not shown in this equation.   

In Figure 1(a) we have the $y_i$'s on a dashed line as an optional input to the MIL block. We will study how the this new feature work in anomaly detection.  

\subsection{Proposed Testing Process}

In the training of BERT and the downstream MIL block, the input features are always segmented into 32 segments. This is required by both the standard implementation of the BERT and the MIL block. However, this is not required in the testing or inference mode.  

In Figure 1(b) a testing pipeline is shown, where the video classification score $p(\hat{y}_{cls})$ is combine with the MIL snippet scores $s(v_i)$. So the final snippet anomaly score is, 

\begin{equation}
    score(v_i) = s(v_i) p(\hat{y}_{cls}).    \label{eq5}
\end{equation}
This is called a score correction method in \cite{AAAI22}. We conceive this idea before we find the work in \cite{AAAI22}. However, \cite{AAAI22} uses the feature after the transformer, while in our work, we use the original feature in the MIL block.  

In Figure 1(b) we use the original feature $f_i$ as input to the MIL block in an online mode, where features can go into the MIL block for processing as they become available. We can also use $x_i$ in an offline mode, where all the video features are divided into 32 segments before they go into the MIL block. This online mode is used in many previous work's implementation. However, the BERT block requires to have 32 segment features, so Figure 1(b) can only actually work in an offline mode. 

In Figure 1(c) we show a simplified testing mode, where the video classification score is not used. In this mode, the BERT is used in the training process but is not used at all at testing time. This make this model very attractive since it has very low complexity. 

\subsection{Combining with RTFM}
The proposed method can work alone, or combine with other anomaly detection methods. We use the RTFM \cite{RTFM} as an example. In this case, the MIL ranking loss is replaced with the feature magnitude based ranking loss of RTFM, and our BCE loss is balanced with RTFM's BCE loss. So the total loss function is defined as,  

\begin{equation}
    l = \beta BCE_{BERT} + (1-\beta) BCE_{RTFM} + RTFM ranking   
    \label{eq6}
\end{equation}
where the BERT BCE loss is defined in last line of Equation (4), and the RTFM BCE loss and ranking loass can be found in \cite{RTFM}. This proposed solution is called RTFM-BERT.  

In RTFM, the BCE loss is defined on the top-k snippets whose feature magnitudes are the largest in a video. They call this classifier a snippet classifier. Since these top-k snippets are selected per video, and their anomaly scores are averaged to represent the video, this score actually represent the whole video class. After doing some analysis we find that this is one of the key contributions of RTFM. When we try to remove this BCE loss, or replace with a snippet score ranking loss as in \cite{UCF-Crime}, the performance of RTFM becomes a lot worse. So the authors of RTFM may not realize that they have put the power of video classification implicitly in their solution.   

\subsection{Discussion: Why is video classification important} 

From Equation (5) we can see that the video classification score ($p(\hat{y}_{cls})$) helps the snippet prediction score $s(v_i)$. For an anomalous snippet, the video classification score does not have effect since if it is not used, then the snippet score is simply $s(v_i)$. For a normal snippet, when the video classification score is small (near 0), then the snippet prediction score is suppressed further smaller. This helps to reduce the chance that a normal snippet is mistakenly classified as an anomalous one. 

In addition, even if the video classification score is not used in Equation (5), the video classification also helps the anomaly detection implicitly. In Equation (4), the video classification is on the BERT output classification vector $\hat{y}_{cls}$, which is a function of the input features $v_i$. The video classification is correct if the MIL selects the correct max-scored instance in the anomalous and normal bags. So this explicit video classification helps the MIL to select the correct instances.

\section{Experiments}

\subsection{Datasets}

We use three anomaly detection datasets: UCF-Crime \cite{UCF-Crime}, ShanghaiTech \cite{shanghaitech}, and newly released XDViolence \cite{XD-Violence}. We do most of our ablation study on UCF-Crime.  

UCF-Crime \cite{UCF-Crime} is a large-scale anomaly detection dataset that contains 1900 untrimmed videos with a total duration of 128 hours from real-world street and indoor surveillance cameras. UCF-Crime consists of complicated and diverse backgrounds. Both training and testing sets contain the same number of normal and abnormal videos. The data set covers 13 classes of anomalies in 1,610 training videos with video-level labels and 290 test videos with frame-level
labels.

ShanghaiTech \cite{shanghaitech} is a medium-scale dataset from a fixed street video surveillance camera. It has 13 different background scenes and 437 videos, including 307 normal videos and 130 anomaly videos. The original dataset is a popular benchmark for the anomaly detection task that assumes the availability of normal training data. Zhong et al. \cite{adgcn_cvpr19} reorganised the data set by selecting a subset of anomalous testing videos into training data to build a weakly supervised training set, so that both training and testing sets cover all 13 background scenes. 

XD-Violence \cite{XD-Violence} is a recently proposed large-scale multiscene anomaly detection data set, collected from real life movies, online videos, sport streaming, surveillance cameras and CCTVs. The total duration of this data set is over 217 hours, containing 4754 untrimmed videos with video-level labels in the training set and frame-level labels in the testing set. It is currently the largest publicly available video anomaly detection data set. 

\subsection{Evaluation Metrics}

We follow previous work \cite{UCF-Crime}, \cite{RTFM}, \cite{AAAI22} to use the  frame-level area under the ROC curve (AUC) as the evaluation metrics on the UCF-Crime and the Shanghai Tech datasets. Following \cite{XD-Violence}, we use average precision (AP) as the evaluation metric on the the XD-Violence dataset. 

Please note that there are two ways to evaluate the AUC or AP performance on the testing dataset. Since the video features are first in snippet, then divided to 32 segments, there are also two ways to do the testing, where the snippet or segment features are used as input to the MIL block, and their performance differs. We use the better of these two to benchmark previous and our work, unless specified otherwise. In the first way where snippet feature is used, the MIL score is mapped back to frame simply by repeating the score 16 times. In the second way where the segment feature is used, we first do an inverse segmentation $seg_{-1}$ as in Equation (3) where each segment's score is mapped back to the original snippets, then every snippet's score is repeated 16 times. 

\subsection{Implementation Details}

We implement the BERT and MIL in PyTorch, the BERT code is borrowed from \cite{BERT}. The default BERT uses 2 layers and 8 attention heads. An initial learning rate of 1E-4 is used, and the training runs 100 epochs. After that some manual fine tune may be used. Two dataset iterators, one for the abnormal data and the other for the normal data, are used. This way, the pairing of abnormal and normal data is random, even when the numbers of abnormal and normal samples are different. We use the Adam optimizer. 

For the video snippet feature, we use two different I3D network. In \cite{RTFM}, the  I3D with Resnet50 \cite{NonLocal2018}, whose feature dimension is 2048, is used. In \cite{XD-Violence} and \cite{park2020learning}, the I3D with Resnet18 \cite{8099985}, whose feature dimension is 1024 is used.  We use two sets of UCF-Crime and ShanghaiTech pre-generated features we find online, one used in \cite{park2020learning} and the other used in in \cite{RTFM}. The first one  has both RGB and optical flow (simply called Flow hereafter) without multiple-crop augmentation,  the other one has RGB only with 10-crop augmentation. On the UCF-Crime, we call the first feature set UCF-Crime, and the second on UCF-Crime-RTFM. For the XD-Violence dataset, we use the 5-crop augmented RGB feature generated by the author of the dataset \cite{XD-Violence}.  

For the evaluation of the combination of our BERT video classification on top of RTFM, we use the code base of RTFM and add BERT into it. As an ablation study, we add one FC layer at the input of the snippet feature, when the feature dimension is 2048. With this change the number of total model parameter is reduced significantly while the impact on performance is negligible.

\begin{table}[tb]
\begin{center}
  \begin{tabular}{|l|c|c|l|}
    \hline
    Feature Set  & L2-Norm? & Segment& Snippet \\
      &  & AUC(\%) & AUC(\%) \\
    \hline
    \cite{park2020learning} & Yes & 80.05 & - \\
    \cite{RTFM} & No & 80.03 & 79.22 \\
    \cite{RTFM} & Yes & 79.59 & 79.54\\
  \hline
  \end{tabular}
  \caption{Prestudy results of two UCF-Crime feature sets. Standard MIL and RGB modality is used. One-crop is used in all cases. 32 segments are used in the testing mode.}
  \label{T1}
  \end{center}
  \end{table}

\begin{table}[tb]
\begin{center}
  \begin{tabular}{|l|c|c|l|}
    \hline
    Dataset & Modality &  AUC(\%) & AUC-2(\%) \\

    \hline
    UCF-Crime & RGB &  82.69 & 79.76 \\
    UCF-Crime & Flow &  85.56 & 80.64\\
    UCF-Crime & RGB+Flow & 86.71 & 82.34 \\
    \hline
    ShanghaiTech & RGB &   91.55 & 80.59\\
    ShanghaiTech & Flow &  96.75 & 83.31\\
    ShanghaiTech & RGB+Flow &  97.54 & 87.62\\

  \hline
  \end{tabular}
  \caption{BERT-MIL performance on UCF-Crime and ShanghaiTech}
  \label{T2}
  \end{center}
  \end{table}

\subsection{Pre-study on Feature Sets}

Since we use two sets of pre-generated feature sets on UCF-Crime and ShanghaiTech, we first want to confirm the consistency of these two feature sets. In addition to different modality and crop, the first feature set is L2 normalized, while the second feature set is not.  

In this experiment we test the standard MIL, RGB only, and 1-crop only. The feature set \cite{park2020learning} is already L2-normalized and divided into 32 segments. For the RTFM \cite{RTFM} feature set, we test both L2-norm and non-L2-norm. In RTFM design, L2 is not used, while L2 normalization is suggested in original MIL work \cite{UCF-Crime}. We use the 32-segments in the testing mode. The results are listed in Table 1. We see that the results are all very close to each other. On the RTFM feature set, non- L2-norm performs slightly better. So in our following comparison, we always use the best result of every method. From this prestudy, we see that these two different feature sets give about same performance. 

\subsection{MIL-BERT on UCF-Crime and ShanghaiTech}

In this study we use the UCF-Crime feature set pre-generated in \cite{park2020learning}. We use 32 segments in the testing mode. These features do not use multiple cropping, and are L2-normalized. 

We first check the video classification only to look at how accurate it is. For this purpose all MIL functions are turned off.  The accuracy results for the RGB, Flow, and RGB+Flow are 83.45\%,85.52\%, 90.00\%, respectively. 

After that we turn the MIl functions back on, and the MIL-BERT is trained end to end. For ablation study purpose, we test a two-step training: first train the video classifier, then freeze BERT and train MIL only. The performance is same as the end to end training. The experiment results are listed in Table 2. Note that in this table, AUC uses the segment score defined in Equation (5), while AUC-2 do not use the video classification score $p(\hat{y}_{cls}$) in Equation (5). The AUC and AUC-2 represent the testing mode in Figure 1(b) and 1(c), respectively. 
We find that the RGB is the weakest modality in UCF-Crime, the Flow is better, and RGB+Flow gives the best performance. We observe that our best AUC with RGB+Flow modality is better than the SOTA result. 

We repeat the test on ShanghaiTech. Again we use the feature set pre-generated in \cite{park2020learning}. We use 32 segments in the testing mode, one-crop, and L2-normalized features. The experiment results are listed in Table 2. The trend is very similar as on UCF-Crime. The RGB+Flow gives the best AUC of 97.54\%, exceeding the SOTA result already.  

\begin{table}[tb]
\begin{center}
  \begin{tabular}{|l|c|c|c|l|}
    \hline
    Dataset & FC? & $\beta$ & AP or &AP-2 or \\
    & & & AUC (\%) & AUC-2 (\%) \\ 
    \hline
    XD-Violence & No & 0.5 & 82.10 & 77.77 \\
    XD-Violence & No & 0.7 & 79.00 & 74.40 \\
    \hline
    UCF-Crime & No & 0.5 & 84.33 & 83.80 \\
    UCF-Crime & No & 0.7 & 83.84 & 83.08 \\
    UCF-Crime & Yes & 0.5 & 84.12 & 83.50 \\
  \hline
  \end{tabular}
  \caption{RTFM-BERT on UCF-Crime and XD-Violence. RGB, 5-crop, and snippet feature are used. FC denotes a FC layer is used on input feature to reduce dimension to 1024. On UCF-Crime, the metric is AUC. }
  \label{T3}
  \end{center}
  \end{table}

\begin{table}[tb]
\begin{center}
  \begin{tabular}{|l|l|l|}
    \hline
    Classifier & Feature & AUC (\%) \\
    \hline
    BERT &  $x_i$ & 86.71 \\
    BERT &  $y_i$ & 83.75 \\
    \hline
    LSTM & $x_i$ & 86.59\\
  \hline
  \end{tabular}
  \caption{Performance of using BERT features $y_i$ in MIL-BERT on UCF-Crime RGB+Flow modality. Also included is an LSTM based video classifier with feature $x_i$.}
  \label{T5}
  \end{center}
  \end{table}

\begin{table*}[tb]
\begin{center}
  \begin{tabular}{|l|c|c|c|l|}
    \hline
    Dataset & Method & Feature & Crop & AUC or AP(\%) \\
    \hline
    UCF-Crime & \cite{adgcn_cvpr19} & I3D-RGB & 10 & 82.12 \\
    UCF-Crime & RTFM\cite{RTFM} & I3D-RGB & 10 & 84.30 \\
    UCF-Crime & \cite{CRFD} & I3D-RGB & 10 & 84.30 \\
    UCF-Crime & \cite{wsal_tip21} & I3D-RGB & 10 & 85.38 \\ 
    UCF-Crime & \cite{AAAI22} & VideoSwin-RGB & 10 & 85.62 \\ 
    UCF-Crime & Ours & I3D-RGB+Flow & 1 & \textbf{86.71} \\ 
    \hline
    ShanghaiTech & RTFM\cite{RTFM} & I3D-RGB & 10 & 97.21 \\
    ShanghaiTech & \cite{wsal_tip21} & I3D-RGB & 10 & 85.30 \\ 
    ShanghaiTech & \cite{MIST} & I3D-RGB & 1 & 94.83 \\ 
    ShanghaiTech & \cite{CRFD} & I3D-RGB & 1 & 97.48 \\ 
    ShanghaiTech & \cite{AAAI22} & VideoSwin-RGB & 10 & 97.32 \\
    ShanghaiTech & Ours & I3D-RGB+Flow & 1 & \textbf{97.54} \\ 
    \hline
    XD-Violence & RTFM\cite{RTFM} & I3D-RGB & 5 & 77.81 \\
    XD-Violence & \cite{XD-Violence} & I3D-RGB+audio & 5 & 78.64 \\ 
    XD-Violence & \cite{CRFD} & I3D-RGB & 1 & 75.90 \\ 
    XD-Violence & \cite{AAAI22} & VideoSwin-RGB & 5 & 78.59 \\
    XD-Violence & Ours & I3D-RGB & 5 & \textbf{82.10} \\ 
    
  \hline
  \end{tabular}
  \caption{Comparison with SOTA results on three datasets.}
  \label{T6}
  \end{center}
  \end{table*}

\subsection{RTFM-BERT on UCF-Crime and XD-Violence}

We choose to test RTFM-BERT on UCF-Crime \cite{UCF-Crime} and XD-Violence \cite{XD-Violence}. We add the BERT video classification into the RTFM \cite{RTFM}. As one ablation study, we add a FC layer at the input of RTFM to reduce the feature dimension from 2048 to 1024 on the UCF-Crime. The dimension of XD-Violence is 1024 already.   

As in RTFM \cite{RTFM}, we only use RGB modality, 10-crop, non-L2-normalized pre-generated feature set on UCF-Crime and RGB modality, 5-crop, non-L2-normalized pre-generated feature set on XD-Violence. Snippet features are used in the testing mode. We follow the same implementation details of RTFM. The results are listed in Table 3. Note that in this table, AP and AUC uses the snippet score defined in Equation (5), while AP-2 and AUC-2 do not use the video classification score $p(\hat{y}_{cls}$) in Equation (5). The AP-2 and AUC-2 represent the testing mode in Figure 1(c). 

From the results, we observe that the BERT video classification does not bring much performance improvement on UCF-Crime. Furthermore, the two APs are very close to each other. $\beta=0.5$ gives better results than $\beta=0.7$.     

However, on the XD-Violence dataset, the BERT video classification brings surprising performance improvement. This is perhaps due to different attribute of dataset and is left for further study.  

\subsection{MIL Using BERT Features} 

Similar to \cite{AAAI22}, where a transformer is used for both video classification and feature refinement, we test using BERT refined feature in MIL block.  We use the same setting as in MIL-BERT on UCF-Crime using RGB+Flow modality. The result is listed in Table 4. From the results we notice that the performance of using $y_i$ is much worse. So this feature is not refined, but deteriorated. No need to mention that, we want an online testing mode shown in Figure 1(c) that offers a low complexity solution. The authors of \cite{AAAI22} might not have realized this effect. We predict that if they implement the transformer as a video classifier only, their performance might be even better.      

\subsection{LSTM Video Classifier} 
In the previous experiments, we also demonstrate that the transformer based video classifier works very closely to the BERT video classifier. As an ablation study, we implement a simpler LSTM based video classifier with two layers and hidden layer dimenstion same as input dimension. Only the feature $x_i$ is used in this classifier. The result is put in the third panel of Table 3. From the result we observe that its performance is almost identical to the BERT classifier, even though its complexity is a  lot smaller (about 1/4) than the BERT. This proves the key contribution of this work, as the title states, the overlooked power of video classification is now implemented.  

\subsection{Comparison with SOTA}  

We compare our best results on UCF-Crime, ShanghaiTech and XD-Violence with SOTA results in the literaturem as shown in Table 5. We show our best AUC or AP result on each dataset. Note that we only include best results published in recent a few years. Older results a lot worse than the SOTA results, including those of the unsupervised ones, are not included. 

From the results, we achieve new SOTA results on all three datasets. Please note that on UCF-Crime and ShanghaiTech, we use RGB+Flow modality while many previous works use RGB modality only \cite{RTFM}, \cite{AAAI22}. On the XD-Violence, we use RGB modality and achieve AP \%82.10, a 3.51\% jump from previous \%78.59 \cite{XD-Violence} which uses both video and audio signals.   

\begin{figure*}[tbh]
    \centering
    \includegraphics[scale=0.475]{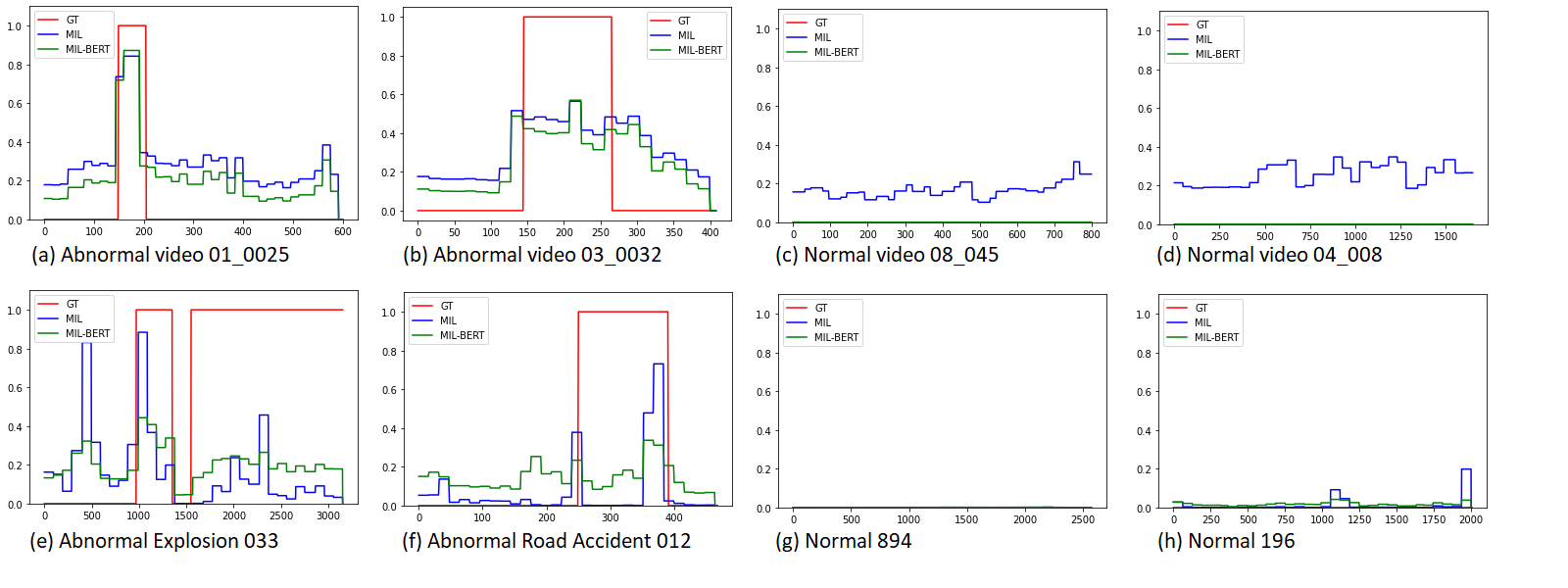}
    \caption{Visualization of anomaly score curves. The videos in the first row are from ShanghaiTech, and videos in the second row are from the UCF-Crime.} 
    \label{fig2}
\end{figure*}

\subsection{Qualitative Analysis}  
Anomaly score curves of some examples videos are shown Figure 2. We see that the BERT video classification work very effectively to press down scores for normal snippets. It may improve or degrade the scores for anomalous snippets. With this video classification, the decision threshold is lowered. The overall impact is reflected in the AUC values. 

\section{Conclusion}

Previously in video anomaly detection with the MIL frameworks, almost all of them overlook or do not realize enough the power of the video classification.  

In this paper, we study the effect of the video classification explicitly.  We propose an video classification using BERT or LSTM. This single change brings significant performance gain. On the RGB+Flow on the UCF-Crime and ShanghaiTech datasets, our proposed MIL-BERT achieves ROC AUC exceeding SOTA results. On the XD-Violence dataset, our proposed RTFM-BERT achieves AP exceeding SOTA resutls by 3.51\%. These experiment results demonstrate the power of video classification. It can be combined with other anomaly detection algorithms to get the best performance. 

%%%%%%%%% REFERENCES
{\small
\bibliographystyle{ieee_fullname}
\bibliography{egbib}

\begin{thebibliography}{10}\itemsep=-1pt

\bibitem{DETR}
Nicolas Carion, Francisco Massa, Gabriel Synnaeve, Nicolas Usunier, Alexander
  Kirillov, and Sergey Zagoruyko.
\newblock End-to-end object detection with transformers.
\newblock {\em ECCV}, 2020.

\bibitem{8099985}
João Carreira and Andrew Zisserman.
\newblock Quo vadis, action recognition? a new model and the kinetics dataset.
\newblock In {\em CVPR}, pages 4724--4733, 2017.

\bibitem{I3D}
Joao Carreira and Andrew Zissermana.
\newblock Quo vadis, action recognition? a new model and the kinetics dataset.
\newblock {\em CVPR}, 2017.

\bibitem{BERT-0}
Jacob Devlin, Ming-Wei Chang, Kenton Lee, and Kristina Toutanova.
\newblock Bert: Pre-training of deep bidirectional transformers for language
  understanding.
\newblock {\em arXiv}, 2018.

\bibitem{ViT}
Alexey Dosovitskiy, Lucas Beyer, Alexander Kolesnikov, Dirk Weissenborn,
  Xiaohua Zhai, Thomas Unterthiner, Mostafa Dehghani, Matthias Minderer, Georg
  Heigold, Sylvain Gelly, Jakob Uszkoreit, and Neil Houlsby.
\newblock An image is worth 16x16 words: Transformers for image recognition at
  scale.
\newblock {\em ICLR}, 2021.

\bibitem{X3D}
Christoph Feichtenhofer.
\newblock X3d: Expanding architectures for efficient video recognition.
\newblock {\em CVPR}, 2020.

\bibitem{MIST}
Jia-Chang Feng, Fa-Ting Hong, and Wei-Shi Zheng.
\newblock Mist: Multiple instance self-training framework for video anomaly
  detection.
\newblock {\em CVPR}, 2021.

\bibitem{video_transformer}
Rohit Girdhar, João Carreira, Carl Doersch, and Andrew Zisserman.
\newblock Video action transformer network.
\newblock {\em CVPR}, 2018.

\bibitem{BERT}
M~Esat Kalfaoglu, Sinan Kalkan, and A~Aydin Alatan.
\newblock Late temporal modeling in 3d cnn architectures with bert for action
  recognition.
\newblock In {\em ECCV Workshops}, pages 731--747. Springer, 2020.

\bibitem{jimaging4020036}
B.~Ravi Kiran, Dilip~Mathew Thomas, and Ranjith Parakkal.
\newblock An overview of deep learning based methods for unsupervised and
  semi-supervised anomaly detection in videos.
\newblock {\em Journal of Imaging}, 4(2), 2018.

\bibitem{MoViNet}
Dan Kondratyuk, Liangzhe Yuan, Yandong Li, Li Zhang, Mingxing Tan, Matthew
  Brown, and Boqing Gong.
\newblock Movinets: Mobile video networks for efficient video recognition.
\newblock {\em CVPR}, 2021.

\bibitem{JHMDB}
H. Kuehne, H. Jhuang, E. Garrote, T. Poggio, and T. Serre.
\newblock Hmdb: A large video database for human motion recognition.
\newblock In {\em International Conf. on Computer Vision (ICCV)}, 2011.

\bibitem{localizanomaly}
F. Landi, C.~G.~M. Snoek, and R. Gucchiara.
\newblock Anomaly locality in video surveillance.
\newblock {\em arXiv preprint 1901.10364}, 2019.

\bibitem{AAAI22}
Shuo Li, Fang Liu, and Licheng Jiao.
\newblock Self-training multi-sequence learning with transformer for weakly
  supervised video anomaly detection.
\newblock {\em AAAI}, 2022.

\bibitem{MMM19}
Kun Liu and Huadong Ma.
\newblock Exploring background-bias for anomaly detection in surveillance
  videos.
\newblock {\em ACM MM}, 2019.

\bibitem{shanghaitech}
W. Liu, D.~Lian W.~Luo, and S. Gao.
\newblock Future frame prediction for anomaly detection -- a new baseline.
\newblock In {\em 2018 IEEE Conference on Computer Vision and Pattern
  Recognition (CVPR)}, 2018.

\bibitem{wsal_tip21}
Hui Lv, Chuanwei Zhou, Zhen Cui, Chunyan Xu, Yong Li, and Jian Yang.
\newblock Localizing anomalies from weakly-labeled videos.
\newblock {\em IEEE Transactions on Image Processing (TIP)}, 2021.

\bibitem{anomaly-review}
Rashmiranjan Nayak, Umesh~Chandra Pati, and Santos~Kumar Das.
\newblock A comprehensive review on deep learning-based methods for video
  anomaly detection.
\newblock {\em Image and Vision Computing}, 106:104078, 2021.

\bibitem{park2020learning}
Jaeyoo Park, Junha Kim, and Bohyoung Han.
\newblock Learning to adapt to unseen abnormal activities under weak
  supervision.
\newblock In {\em Asian Conference on Computer Vision (ACCV)}, 2020.

\bibitem{UCF101}
Khurram Soomro, Amir~Roshan Zamir, and Mubarak Shah.
\newblock Ucf101: A dataset of 101 human action classes from videos in the
  wild.
\newblock {\em CRCV-TR-12-01}, 2012.

\bibitem{UCF-Crime}
W. Sultani1, C. Chen, and M. Shah.
\newblock Real-world anomaly detection in surveillance videos.
\newblock In {\em CVPR}, 2018.

\bibitem{RTFM}
Y. Tian, G. Pang, Y. Chen, R. Singh, J. Verjans, and G. Carneiro.
\newblock Weakly-supervised video anomaly detection with robust temporal
  feature magnitude learning.
\newblock {\em arXiv preprint arXiv:2101.10030}, 2021.

\bibitem{C3D}
Du Tran, Lubomir Bourdev, Rob Fergus, Lorenzo Torresani, and Manohar Paluri.
\newblock Learning spatiotemporal features with 3d convolutional networks.
\newblock {\em ICCV}, 2015.

\bibitem{attention}
Ashish Vaswani, Noam Shazeer, Niki Parmar, Jakob Uszkoreit, Llion Jones,
  Aidan~N. Gomez, Lukasz Kaiser, and Illia Polosukhin.
\newblock Attention is all you need, 2017.

\bibitem{NonLocal2018}
Xiaolong Wang, Ross Girshick, Abhinav Gupta, and Kaiming He.
\newblock Non-local neural networks.
\newblock {\em CVPR}, 2018.

\bibitem{STAD}
Jie Wu, Wei Zhang, Guanbin Li, Wenhao Wu, Xiao Tan, Yingying Li, Errui Ding,
  and Liang Lin.
\newblock Weakly-supervised spatio-temporal anomaly detection in surveillance
  video.
\newblock {\em IJCAI}, 2019.

\bibitem{IJCAI21}
Jie Wu, Wei Zhang, Guanbin Li, Wenhao Wu, Xiao Tan, Yingying Li, Errui Ding,
  and Liang Lin.
\newblock Weakly-supervised spatio-temporal anomaly detection in surveillance
  video.
\newblock {\em IJCAI}, 2021.

\bibitem{CRFD}
P. Wu and J. Liu.
\newblock Learning causal temporal relation and feature discrimination for
  anomaly detection.
\newblock {\em IEEE Transactions on Image Processing}, 30:3513--3527, 2021.

\bibitem{XD-Violence}
Peng Wu, jing Liu, Yujia Shi, Yujia Sun, Fangtao Shao, Zhaoyang Wu, and Zhiwei
  Yang.
\newblock Not only look, but also listen: Learning multimodal violence
  detection under weak supervision.
\newblock In {\em European Conference on Computer Vision (ECCV)}, 2020.

\bibitem{adgcn_cvpr19}
Jia-Xing Zhong, Nannan Li, Weijie Kong, Shan Liu, Thomas~H. Li, and Ge Li.
\newblock Graph convolutional label noise cleaner: Train a plug-and-play action
  classifier for anomaly detection.
\newblock In {\em CVPR}, 2019.

\end{thebibliography}
}

\end{document}